\documentclass[10pt,twocolumn,letterpaper]{article}

\usepackage{wacv}
\usepackage{times}
\usepackage{epsfig}
\usepackage{graphicx}
\usepackage{amsmath}
\usepackage{amssymb}
\usepackage{booktabs}

\usepackage{bm}
\usepackage{multirow}
\usepackage[accsupp]{axessibility} 

\newcommand{\new}[1]{#1}

\newcommand{\norm}[1]{\left\lVert#1\right\rVert}

\def\wacvPaperID{147} 

\wacvalgorithmstrack   

\wacvfinalcopy 

\ifwacvfinal
\usepackage[breaklinks=true,bookmarks=false,colorlinks]{hyperref}
\else
\usepackage[pagebackref=true,breaklinks=true,colorlinks,bookmarks=false]{hyperref}
\fi

\begin{document}

\title{Progressive Video Summarization via Multimodal Self-supervised Learning}

\author{Haopeng Li\textsuperscript{1}, Qiuhong Ke\textsuperscript{3}, Mingming Gong\textsuperscript{2}, Tom Drummond\textsuperscript{1}\\
\textsuperscript{1}School of Computing and Information Systems, The University of Melbourne\\
\textsuperscript{2}School of Mathematics and Statistics, The University of Melbourne\\
\textsuperscript{3}Department of Data Science \& AI, Monash University\\
{\tt\small haopeng.li@student.unimelb.edu.au, \{mingming.gong, tom.drummond\}@unimelb.edu.au,}\\
{\tt\small qiuhong.ke@monash.edu}
}

\maketitle
\thispagestyle{empty}

\begin{abstract}
   Modern video summarization methods are based on deep neural networks that require a large amount of annotated data for training. However, existing datasets for video summarization are small-scale, easily leading to over-fitting of the deep models. Considering that the annotation of large-scale datasets is time-consuming, we propose a multimodal self-supervised learning framework to obtain semantic representations of videos, which benefits the video summarization task. Specifically, the self-supervised learning is conducted by exploring the  semantic consistency between  the videos and text in both coarse-grained and fine-grained fashions, as well as recovering  masked frames in the  videos. The multimodal framework is trained on a newly-collected dataset that consists of video-text pairs. Additionally, we introduce a progressive video summarization method, where the important content in a video is pinpointed progressively to generate better summaries. Extensive experiments have proved the effectiveness and superiority of our method in rank correlation coefficients and F-score\footnote{The codes and dataset will be released soon.}.
\end{abstract}

\section{Introduction}
\label{sec:intro}

Video summarization aims to generate a short version of a video by picking the most important frames or shots containing the main content of the original video, which greatly improves the efficiency of video browsing and retrieval. State-of-the-art video summarization methods are based on deep neural networks which model the dependencies between frames/shots and estimate their importance  
\cite{zhao2021reconstructive,jung2020global,jung2019discriminative,rochan2018video,zhang2018retrospective,he2019unsupervised}. However, existing datasets for video summarization are relatively small \cite{gygli2014creating,song2015tvsum}, which easily leads to over-fitting of the deep models. Meanwhile, collecting a large-scale annotated dataset for video summarization is challenging and time-consuming, as multiple annotators need to provide frame/shot-level annotations to minimize subjectivity. When the annotated data are scarce, self-supervised
learning has shown great power in boosting the performance
of deep models in various scenarios, such as image retrieval \cite{noroozi2016unsupervised}, action recognition \cite{ahsan2019video}, and language understanding \cite{devlin2018bert}. Encouraged by these successful stories, one may ask a natural question, ``\textbf{Can self-supervised learning benefit video summarization?}"


\begin{figure}[tbp]
\centering
\includegraphics[width=0.95\columnwidth]{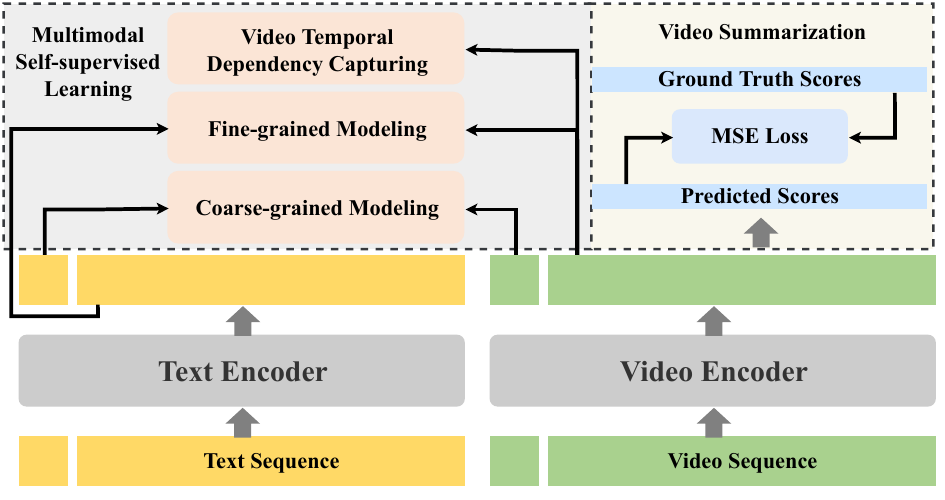}
\caption{\new{The proposed multimodal self-supervised framework for video summarization.}}
\label{abstract}
\end{figure}

\new{In this paper, we show that the answer to the above question is \textbf{YES}. Inspired by the semantic correlations between videos and their text information, we propose a 
new
multimodal self-supervised framework for video summarization. In this framework, the multimodal correlations are captured from two aspects: 1) The coarse-grained modeling uses sequence-level representations of video and text to predict whether they have semantic correspondence; 2) The fine-grained modeling regards the video and the text as two sets and measures their distance using individual words and frames. Meanwhile, we also try to capture the temporal dependencies in the visual modality by modeling the relationship between the masked frame and the video. The proposed  multimodal self-supervised framework is shown in Figure \ref{abstract}.}
To train our multimodal encoder, we collect a new dataset
that consists of video-text pairs.
Specifically, we first obtain several video \textit{categories} and \textit{search queries} from Google Trend\footnote{\url{https://trends.google.com/trends}}. We then collect the top searched videos whose duration is 3--20 
minutes as well as their corresponding \textit{titles} and \textit{descriptions}. Finally, a dataset consisting of 3,081 \textbf{Y}ou\textbf{T}ube \textbf{V}ideo-\textbf{T}ext pairs (\textbf{YTVT}) is 
collected 
for the multimodal self-supervised learning.


After the self-supervised pretraining, a multimodal sequence encoder is obtained and further fine-tuned for the 
video summarization task.
Existing video summarization methods \cite{zhao2021reconstructive,jung2020global,jung2019discriminative,he2019unsupervised} are mainly based on 
a single-stage fashion where the videos are examined only once to generate the final summaries, which might be insufficient to pinpoint the important content.
In this 
work,
we propose a progressive video summarization method using the pretrained multimodal encoder, where the input sequence is refined by iteratively emphasizing the important content in a multi-stage fashion. 
Besides, we also describe how to incorporate the text information 
for video summarization.

Our contributions are summarized as follows:

\noindent1) \new{We introduce multimodal self-supervised learning, where the multimodal correlations are modeled in both coarse-grained and fine-grained fashions. Meanwhile, the temporal dependencies in videos are captured by modeling the relationship between the masked frame and the whole video.}

\noindent2) \new{We collect a dataset of YouTube video-text pairs for multimodal self-supervised learning. The text of each video includes four types of information, depicting the video from the general category to the specific description.}

\noindent3) \new{Based on the pretrained video encoder, we propose a progressive video summarization method where 
the input video sequence 
is enhanced in a multi-stage fashion. 
We also incorporate text information for better video summarization.}

\section{Related Work}

\subsection{Video Summarization}

Video summarization methods can be roughly classified into supervised methods and unsupervised ones. We review existing methods according to the category they belong to.

The unsupervised video summarization \cite{mahasseni2017unsupervised,zhou2018deep,he2019unsupervised,mei20142,mahasseni2017unsupervised} relies on the criteria designed by human, such as representativeness \cite{de2011vsumm,ngo2003automatic} and diversity \cite{zhou2018deep}. Conventional machine learning algorithms such as clustering and dictionary learning were widely exploited in unsupervised methods. For instance, $L_{2,0}$-constrained sparse dictionary learning was used to address video summarization in \cite{mei20142}.
Besides, unsupervised methods based on deep neural networks were presented in recent works. SGAN \cite{mahasseni2017unsupervised} used adversarial generative networks to generate summaries which were hard to discriminate from the original videos.

Most of the supervised methods are based on deep neural networks to model the temporal dependencies \cite{zhang2016video,zhao2017hierarchical,zhao2018hsa,fajtl2018summarizing,zhao2021reconstructive,zhang2018retrospective,rochan2018video,jung2019discriminative}, which requires human summaries for training. Numerous deep models were developed to capture the temporal dependencies in either local fashion or global fashion. For example, long short-term memory (LSTM) was exploited to model the video and predict the frame-level scores in vsLSTM/dppLSTM \cite{zhang2016video}. Furthermore, hierarchical adaptions of LSTM were proposed to address the issues of plain LSTM \cite{zhao2017hierarchical,zhao2018hsa}. Besides, attention models and graph models were exploited to capture the global dependencies. For instance, 
a sequence-graph structure was developed in 
RSGN \cite{zhao2021reconstructive}, which models the frame-level dependencies and the shot-level ones successively. Considering the semantic correlations across videos, VJMHT \cite{9808180} is developed based on a hierarchical Transformer. Nevertheless, most of the existing methods perform video summarization in a single-stage fashion where the videos are examined only once. In contrast, we propose progressive video summarization to iteratively refine the input and pinpoint the important content. Although SumGraph \cite{park2020sumgraph} also uses the recursive idea, our method is different to SumGraph in terms of motivation and methodology. Specifically, SumGraph recursively obtains a graph where the nodes are connected by stories instead of similarities, while our method iteratively emphasizes the important frames by inspecting the videos multiple times. Besides, SumGraph recursively refines the adjacent matrix in GCN, while our method re-weights the input by the scores output from previous stage.

\subsection{Multimodal Self-supervised Learning}

Self-supervised learning has been widely used for the pretraining of deep models, which boosts their performance to a large extent in various fields \cite{fernando2017self,wang2017transitive,jing2020self,zhang2020pegasus,lan2019albert,liu2019roberta,devlin2018bert}. Considering the semantic consistency in multimodal data, contrastive learning were exploited to model the correspondence among different modalities \cite{korbar2018cooperative,li2020oscar,alayrac2020self,arandjelovic2018objects,arandjelovic2017look}. For instance, the consistency of videos and audios were leveraged to train the deep encoder in \cite{korbar2018cooperative}. Besides, the semantic correlations between images and text were used to obtain semantic representations \cite{li2020oscar,sun2019videobert,zhu2020actbert}. Such pretraining was proved effective in many tasks. Furthermore, the consistency among three modalities, video, audio, and text, were considered in \cite{alayrac2020self,akbari2021vatt}, by which a versatile network can be obtained. Self-supervised learning has been exploited for video summarization. Specifically, CLIP-It \cite{narasimhan2021clip} uses the pretrained CLIP to extract frame features and the six-layer Transformer to obtain high performance. But it also brings large computation during testing. However, our method use traditional GoogLeNet features and a three-layer Transformer to achieves significant results (especially in rank-based evaluation) with reasonable computational cost. Besides, for self-supervised learning, we propose a framework that exploits the correspondence between modalities in both coarse-grained and fine-grained fashions, while considering the temporal dependencies in videos.



\section{Multimodal Self-supervised Progressive Video Summarization}

Deep learning has been popularly used for video summarization \cite{jung2020global,zhao2021reconstructive,zhao2018hsa}, yet 
most of the existing  datasets \cite{gygli2014creating,song2015tvsum} are relatively small, resulting in over-fitting of the deep models. To resolve this issue, we explore self-supervised learning to improve video summarization. In this section, we propose multimodal \textbf{S}elf-\textbf{S}upervised \textbf{P}rogressive \textbf{V}ideo \textbf{S}ummarization (\textbf{SSPVS}). Specifically, we first introduce the collected dataset 
for the multimodal self-supervised learning. Then, we elaborate the framework 
for the multimodal self-supervised learning.
We further present the progressive video summarization based on the pretrained encoders. 
Finally, we illustrate how to incorporate text information for better 
video summarization.

\subsection{Dataset for Self-supervised Learning}
\label{dc}
Generally, there exist semantic correlations between videos and their associated text information such as titles and descriptions. Such correlations provide supervision that can be used to train a multimodal network in a self-supervised manner.
This encourages the multimodal network to learn better representations of the video and text, which benefit the 
video summarization task. 
To learn the semantic correlation between the video and text information, video-text data are required. 
To this end, we first collect video data as well as their associated text information. 

\begin{figure}[tbp]
\centering
\includegraphics[width=0.95\columnwidth]{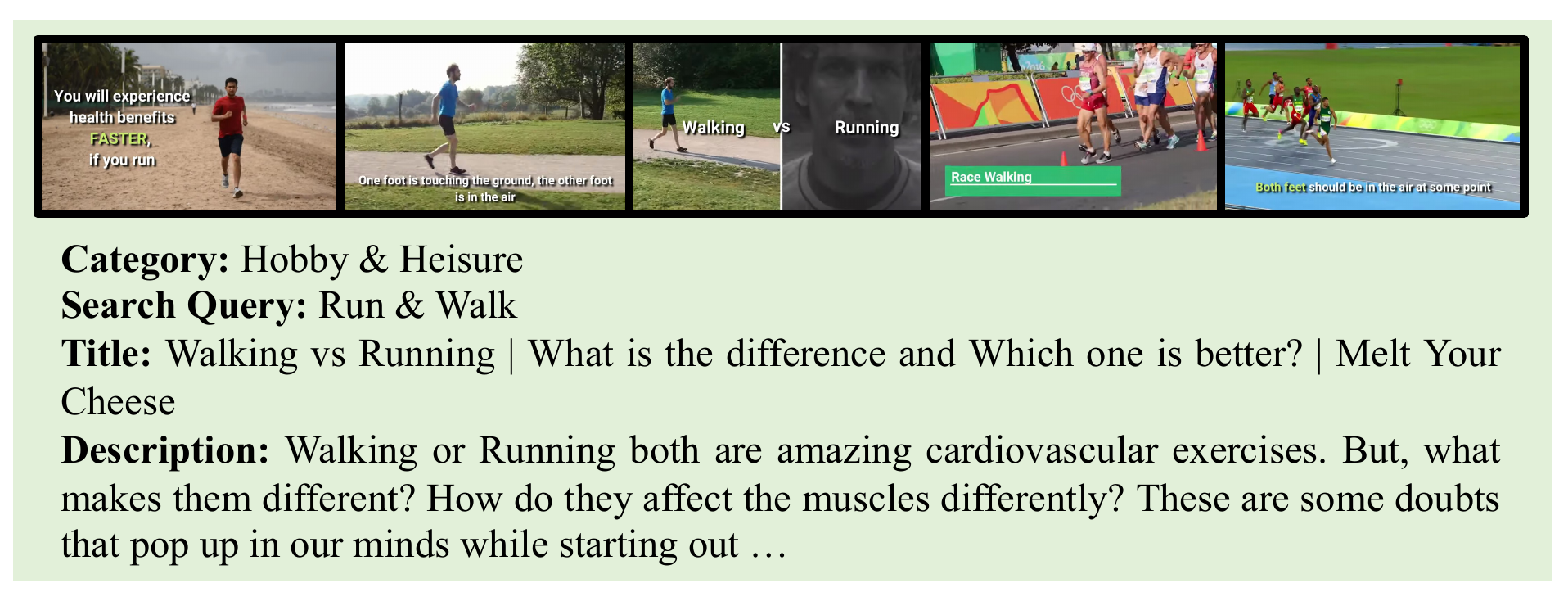}
\caption{An example from the YTVT dataset. Five sampled frames and the text information of 
the video are presented. 
}
\label{data_eg}
\end{figure}

\subsubsection{Data Collection}
In this paper, we collect video data and their associated text information from YouTube. Specifically, we 
first obtain 23 video categories from Google Trend, such as \textit{Autos \& Vehicles} and \textit{Beauty \& Fitness}. For each category, we use its sub-categories as search queries and obtain search results on YouTube. For instance, the category of \textit{Hobby \& Leisure} has sub-categories such as \textit{Cycling} and \textit{Bowling}, and these sub-categories are used as search queries to collect more specific videos.
Note that we manually eliminate trending queries such as \textit{Gossip} and \textit{Celebrity} because such videos contain few general scenarios.
To make the self-supervised model robust to complex multimodal semantic correlations, we collect only the videos longer than 3 minutes to guarantee enough visual diversity of the data.
Furthermore, considering the GPU memory limit, the videos longer than 20 minutes 
are also ruled out. The search results with required length are collected. Besides categories, we also collect the video-specific text information, 
including titles and descriptions. In summary, four types of text for each video 
are obtained: \textit{category}, \textit{search query}, \textit{title}, and \textit{description}. Figure \ref{data_eg} shows an example from YTVT\footnote{More examples can be found in the supplementary materials.}.

\subsubsection{Data Pre-processing}
The collected text information cannot be directly used for self-supervised learning, because it contains a great deal of noise and irrelevant text which has no semantic meaning, especially in the \textit{descriptions}. In this case, we first perform data cleaning by removing the noisy and meaningless text, including extra spaces, special symbols, non-Unicode characters, URLs, E-mails, etc. By this means, the remaining text is semantically related to the corresponding videos, which can be used for video-text joint modeling. Additionally, following the traditional pre-processing steps in NLP, we apply lemmatization to each word and lowercase all letters. Finally, a \textbf{Y}ou\textbf{T}ube-based dataset of 3,081 \textbf{V}ideo-\textbf{T}ext pairs are collected for multimodal self-supervised pretraining, which is named as \textbf{YTVT}. 
Detailed statistics of YTVT (after pre-processing) are shown in Table \ref{ds}.

\begin{table}[tbp]
\caption{Statistics of YTVT. ``Min/Max/Avg. Duration" represents the minimum/maximum/average duration of videos. ``Avg. Title/Desc. Len." represents the average number of words in 
titles/descriptions.}
\label{ds}
\center
\begin{tabular}{lr|lr}
\hline
Statistics & Result & Statistics & Result \\ \hline
\#Videos & 3,081 & Min Duration & 180s\\
\#Categories & 23 & Max Duration &  1,200s \\
\#Queries & 202 & Avg. Duration & 555.2s \\ 
Avg. Title Len. & 9.7 & Avg. Desc. Len. & 98.6 \\ \hline
\end{tabular}
\end{table}

\subsection{Multimodal Self-supervised Pretraining}
\label{ssp}

\begin{figure*}[tbp]
\centering
\includegraphics[width=0.8\textwidth]{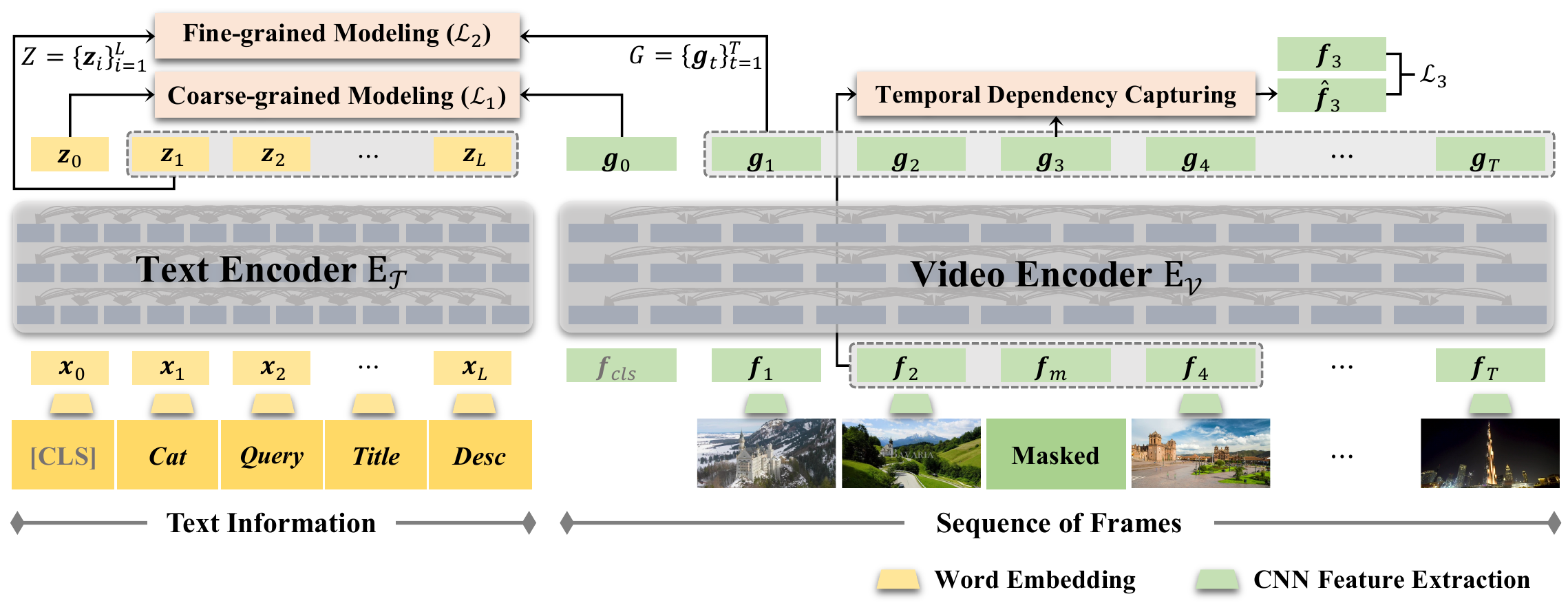}
\caption{\new{The overview of our multimodal self-supervised framework. It consists of a text encoder and a video encoder. The correspondence between the video and the text is modeled in both coarse-grained and fine-grained fashions. Additionally, the temporal dependencies in the video are captured by predicting the masked frame, considering the relationship between the masked frame and the whole video.}
}
\label{ssl}
\end{figure*}


Given the video-text data, we investigate a multimodal network to exploit the correlation between videos and text information and model the temporal dependencies within videos. The framework is shown in Figure \ref{ssl}. Specifically, it 
consists of two unimodal encoders for the text information and the visual information, respectively. We explain the network structure and the learning objectives as follows.

\new{\subsubsection{Network Structure}
\noindent\textbf{Text Encoder.} For the input, four different types of text information of each video are considered, 
i.e., 
\textit{category}, \textit{search query}, \textit{title}, and \textit{description}, each consisting of a sequence of words. The four separate text sequences are combined to form a larger sequence, with a \texttt{[SEP]} token placing between every two types of text sequences to separate each other. The \texttt{[CLS]} token is prepended to the whole sequence to aggregate information from the text. Each word is converted into a vector by the pretrained word embedding module of BERT \cite{devlin2018bert}. The dimension of word embeddings $d_w$ is 768. The pretrained BERT model \cite{devlin2018bert} is adopted to stabilize the text encoding. Formally, the encoded word sequence is denoted as $\left\{\bm{z}_i\right\}_{i=0}^L$, where $\bm{z}_0\in\mathbb{R}^{d_w}$ is the representation of the whole text, $\bm{z}_1,\cdots,\bm{z}_L \in\mathbb{R}^{d_w}$ are the encoded embeddings, and $L$ is the number of words.}

\noindent\new{\textbf{Video Encoder.} Following \cite{zhang2016video,mahasseni2017unsupervised,zhao2018hsa}, the output of the penultimate layer (pool 5) of the GoogLeNet \cite{szegedy2015going} pre-trained on ImageNet \cite{russakovsky2015imagenet} is adopted as the frame feature, which is consistent with most of the video summarization methods for fair comparisons. The dimension of frame features $d_f$ is 1,024. Similar to the \texttt{[CLS]} token for text encoding, a learnable feature ($\bm{f}_{cls}\in \mathbb{R}^{d_f}$) is prepended to the frame features to aggregate the temporal information in the video. The video encoder is a three-layer Transformer with random initialization. Formally, the encoded frame features are denoted as $\left\{\bm{g}_t\right\}_{t=0}^T$, where $\bm{g}_0\in\mathbb{R}^{d_f}$ is the representation of the whole video sequence, $\bm{g}_1,\cdots,\bm{g}_T \in\mathbb{R}^{d_f}$ are the encoded frame features, and $T$ is the number of frames.}

\new{\subsubsection{Learning Objectives}
In this work, we propose two types of learning objectives for self-supervised learning: 1) the semantic correlation between the video and the text, and 2) the temporal dependencies within videos, each of which is explained as follows.}

\noindent\new{\textbf{Cross-modality Semantic Correspondence.} Two sub-objectives are designed to model the correspondence between the video and the text: the coarse-grained modeling and the fine-grained modeling. The coarse-grained modeling exploits sequence-level representations to capture the semantic correlation. For each video-text pair during training, the text information is replaced with that of another video with a probability 50\% to generate the negative pair. Formally, given the representations of the video and the text, $\bm{g}_0,\bm{z}_0$, we use a three-layer perceptron $\mathrm{MLP}_{cls}(\cdot)$ to predict whether the video and the text are corresponding, i.e., $p_{c}=\mathrm{MLP}_{cls}([\bm{g}_0,\bm{z}_0])$,
where $[\cdot,\cdot]$ represents concatenation.
Following previous self-supervised learning \cite{li2020oscar,sun2019videobert}, binary cross entropy is used as the sub-objective,
\begin{equation}
\label{cl}
\mathcal{L}_1=-(y\log(p_{c})+(1-y)\log(1-p_{c})),
\end{equation}
where $y$ is the binary label indicating whether the video-text pair is corresponding or not. The coarse-grained modeling focuses on sequence-level representations which contains only the global information. However, the local information in the video and the text are also of great importance to video understanding. Motivated by this, we propose fine-grained modeling for the correspondence between the video and the text. Formally, given the set of the encoded frames $\mathcal{G}=\left\{\bm{g}_t\right\}_{t=1}^T$ and the set of the encoded word embeddings $\mathcal{Z}=\left\{\bm{z}_i\right\}_{i=1}^L$, we first measure the distance between the two sets using Hausdorff distance $d_H$ as follows,
\begin{equation}
d_H(\mathcal{G},\mathcal{Z})=\max\{d(\mathcal{G},\mathcal{Z}),d(\mathcal{Z},\mathcal{G})\},
\end{equation}
\begin{align}
&d(\mathcal{G},\mathcal{Z})=\max_{t}\min_{i}\norm{\frac{\bm{g}_t}{\norm{\bm{g}_t}_2}-\frac{\mathrm{MLP}_{\mathcal{T}}(\bm{z}_i)}{\norm{\mathrm{MLP}_{\mathcal{T}}(\bm{z}_i)}_2}}_2,\\
&d(\mathcal{Z},\mathcal{G})=\max_{i}\min_{t}\norm{\frac{\bm{g}_t}{\norm{\bm{g}_t}_2}-\frac{\mathrm{MLP}_{\mathcal{T}}(\bm{z}_i)}{\norm{\mathrm{MLP}_{\mathcal{T}}(\bm{z}_i)}_2}}_2,
\end{align}
where we use a two-layer perceptron $\mathrm{MLP}_{\mathcal{T}}(\cdot)$ to map the encoded word embeddings into the visual space. Based on the distance of the two sets, we exploit the contrastive loss to pull the corresponding video-text sets together and push the unmatched sets away, i.e.,
\begin{equation}
\label{l2}
\mathcal{L}_2=yd_H^2+(1-y)\max\{0,m-d_H^2\},
\end{equation}
where $m$ is a pre-defined margin. In this sub-objective, instead of using a holistic representation for the video or the text, we model the semantic correlation between the video and the text by inspecting individual frames and words. By this means, the framework can discover more fine-grained cross-modal information for video understanding.}

\begin{figure*}[tbp]
\centering
\includegraphics[width=0.9\textwidth]{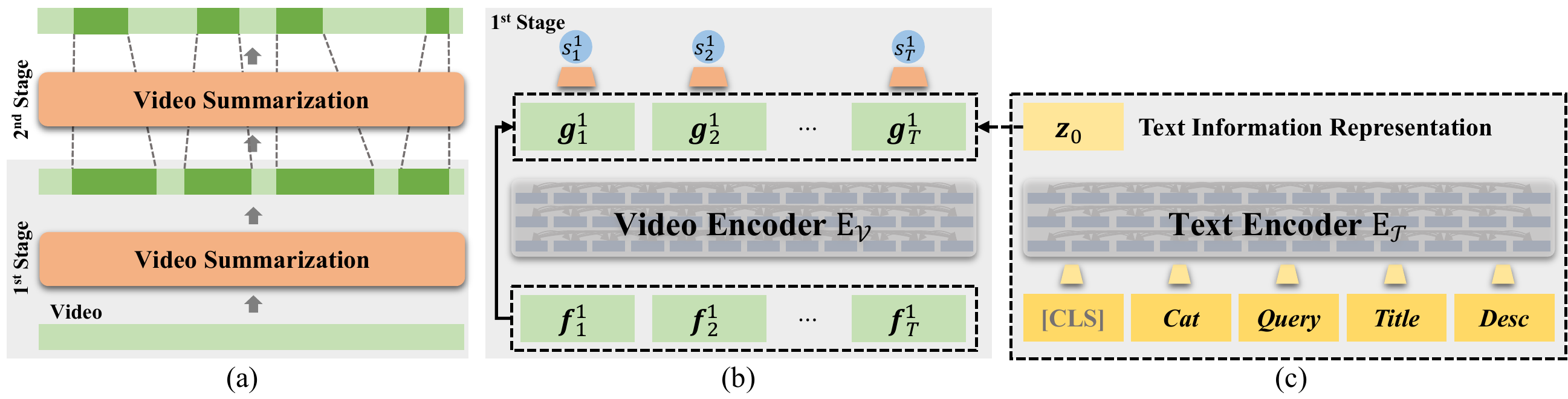}
\caption{\new{(a) The overview of the proposed progressive video summarization of two stages, where the structure and process of each stage is identical. (b) The details of video summarization in the first stage.
(c) The details of the text information encoding. 
} }
\label{pvs}
\end{figure*}

\noindent\new{\textbf{Temporal Dependencies in Videos.} We also capture the temporal dependencies in videos. Similar to the training of BERT, we randomly mask a frame and require the model to recover it. Specifically, we replace a randomly selected frame with a learnable feature ($\bm{m}\in\mathbb{R}^{d_f}$). Instead of predicting the masked frame using the encoded masked feature as most methods, we recover the frame by considering the temporal dependencies between the masked frame and whole video, i.e., whether the masked frame is a smooth transition or an abrupt one. To this end, a two-layer perceptron $\mathrm{MLP}_{s}(\cdot)$ is applied to the encoded masked feature to predict the probability of being a smooth transition. Assuming the $t$-th frame is masked, the probability is computed as $p_{s}=\mathrm{MLP}_s(\bm{g}_{t})$,
where $\bm{g}_{t}$ is the encoded feature of $\bm{m}$. Then, the masked frame is recovered from two aspects: 1) If it is a smooth transition, we recover it by using only its neighbors (local information) with an one-layer Transformer ($\mathrm{T}_{\mathcal{V}}$) and a linear projection as follows,
\begin{equation}
\label{recover1}
\bm{R}=\mathrm{T}_{\mathcal{V}}([\bm{f}_{t-k},\cdots,\bm{m},\cdots,\bm{f}_{t+k}]^{\mathrm{T}}+\bm{E}^\mathcal{V}_{pos}),
\end{equation}
\begin{equation}
\hat{\bm{f}^{1}_t}=\bm{W}_1\bm{R}_c
\end{equation}
where $k$ is a pre-defined window radius, $\bm{E}^\mathcal{V}_{pos}\in\mathbb{R}^{(2k+1)\times {d_f}}$ is the positional encoding, 
$\bm{R}_c\in\mathbb{R}^{{d_f}}$ is the feature in $\bm{R}$ corresponding to $\bm{m}$, and $\bm{W}_1\in\mathbb{R}^{{d_f}\times {d_f}}$ is a learnable parameter. 2) If the masked frame is an abrupt transition (which means only the local information is insufficient for inferring the masked frame), we use $\bm{g}_{t}$ to recover it, as $\bm{g}_{t}$ contains global information of the video. Specifically, a simple linear projection is applied to $\bm{g}_{t}$ to predict the masked frame, i.e., $\hat{\bm{f}^{2}_t}=\bm{W}_2\bm{g}_{t}$,
where $\bm{W}_2\in\mathbb{R}^{{d_f}\times {d_f}}$ is a learnable parameter. Considering the two situations, the masked frame is recovered by combining $\hat{\bm{f}^{1}_t}$ and $\hat{\bm{f}^{2}_t}$, i.e.,
\begin{equation}
\hat{\bm{f}}_t=p_s\hat{\bm{f}}^{1}_t+(1-p_s)\hat{\bm{f}}^{2}_t.
\end{equation}
The loss is defined as the mean squared error between the masked frame and the recovered one, i.e., 
\begin{equation}
\mathcal{L}_3=\frac{1}{d_f}\norm{\hat{\bm{f}_t}-\bm{f}_t}_2^2.
\end{equation}}
Note that there is no label for $p_s$, and we only apply supervision for the recovered frames, which forces the model to adaptively re-weight the predictions from two scenarios.

\new{In summary, the training of the multimodal framework involves three loss functions, and we use their combination for self-supervised learning, i.e.,
\begin{equation}
\label{lssl}
\mathcal{L}_{SSL}=\mathcal{L}_1+\alpha\mathcal{L}_2+\beta\mathcal{L}_3,
\end{equation}
where $\alpha,\beta$ are hyper-parameters to balance the three terms.}

\subsection{Progressive Video Summarization} 
In this section, we describe how to perform summarization using the pretrained encoders.
Existing methods perform video summarization in the single-stage fashion where the videos are examined only once, which 
might be insufficient to pinpoint the important content. To address this limitation, we propose progressive video summarization.
As shown in Figure \ref{pvs}(a), the framework is a stack of multiple models with the identical architecture. Each model is referred to as a \textit{stage} whose structure is shown in Figure \ref{pvs}(b). 

More specifically, the input of the $n$-th stage 
is computed as the weighted enhancement of the input of the previous stage  based on the output of the previous stage,
i.e.,
\begin{equation}
\label{input}
\bm{F}^n=\bm{F}^{n-1}*\bm{s}^{n-1}+\bm{F}^{n-1},
\end{equation}
where $\bm{F}^n=\left[\bm{f}_1^{n},\cdots,\bm{f}_T^{n}\right]^{\mathrm{T}}\in\mathbb{R}^{T\times {d_f}}$ represents the sequence of 
input features at the $n$-th stage. 
$\bm{f}_t^{n} \in\mathbb{R}^{d_f}$ 
denotes the feature at the $t$-th time step of the sequences $\bm{F}^{n}$. 
$\bm{s}^{n-1}= \left[s_1^{n-1},\cdots,s_T^{n-1}\right]^{\mathrm{T}}\in \mathbb{R}^T$ is a sequence of the frame scores output by the 
 $(n-1)$-th 
stage. 
$s_t^{n-1}\in[0,1]$ is a scalar denoting the score at the $t$-th time step of $\bm{s}^{n-1}$.
$T$ represents
the number of frames in the sequence. $*$ represents row-wise multiplication. For the first stage, $\bm{F}^{0}$ is initialized as the original frame features extracted by the pretrained CNN, and $\bm{s}^{0}:=\bm{0}$.

\new{The underlying motivation of formulating $\bm{F}^n$ as in Eq. (\ref{input}) is to iteratively refine the video sequence by emphasizing the important content. We find that, even though the video encoder receives scaled versions of the input sequence in the $n$-th stage ($n>1$) (due to the residual connection), it can still model the sequence properly and performs much better than the formulation without the residual connection (\textbf{see supplementary materials for details}).}

At the $n$-th stage, the input feature $\bm{F}^n$ is encoded by the pretrained video encoder ($\mathrm{E}_{\mathcal{V}}$), 
i.e.,
\begin{equation}
\bm{G}^n=\mathrm{E}_{\mathcal{V}}(\bm{F}^n+\bm{E}^\mathcal{V}_{pos}),
\end{equation}
where $\bm{G}^n=\left[\bm{g}_1^{n},\cdots,\bm{g}_T^{n}\right]^{\mathrm{T}}\in\mathbb{R}^{T\times {d_f}}$ is the sequence of encoded features. 
$\bm{g}_t^{n} \in\mathbb{R}^{d_f}$ denotes the feature at $t$-th time step of $\bm{G}^{n}$.
$\bm{E}^\mathcal{V}_{pos}\in\mathbb{R}^{T\times {d_f}}$ is the positional encoding for the video sequence. 
Note that the video encoders in different stages are identical and share the parameters.
Containing the global temporal information of the sequence, 
$\bm{G}^n$ is then leveraged to predict the frame-level importance scores. Additionally, a residual connection is applied before the linear projection to improve training stability \cite{he2016deep}, i.e.,
\begin{equation}
\bm{s}^n=\sigma ((\bm{G}^n+\bm{F}^n)\bm{W}^n+b^n),
\end{equation}
where $\bm{s}^n\in \mathbb{R}^T$ is the output sequence of the frame scores at 
the $n$-th stage, $\bm{W}^n\in \mathbb{R}^{d_f},b^n\in \mathbb{R} $ are the learnable parameters of the $n$-th stage, and $\sigma(\cdot)$ is the Sigmoid function. 

The final frame scores are computed by taking into consideration 
the score sequences output by 
all stages, i.e.,
\begin{equation}
\bm{s}^*=\bm{s}^1\odot\bm{s}^2\odot\cdots\odot\bm{s}^N\in\mathbb{R}^T,
\end{equation}
where $\odot$ is element-wise multiplication, and $N$ is the number of stages. \textbf{More manners to compute the final scores can be found in the supplementary materials}.

\noindent\textbf{Optimization and Summary Generation.} 
We use the the ground truth frame-level importance scores provided in the datasets to train the proposed video summarization framework.
Formally, given the ground truth frame scores $\bm{s}_{gt}\in\mathbb{R}^T$ and the predicted frame scores $\bm{s}^{*}$ of a video, the mean square error is exploited as the loss function, i.e., 
\begin{equation}
\mathcal{L}_{VS}=\frac{1}{T}\norm{\bm{s}^{*}-\bm{s}_{gt}}_2^2.
\end{equation}

To generate summaries, we follow \cite{zhang2016video,mahasseni2017unsupervised,zhou2018deep} and 
select shots to maximize the total score, with the constraint that the length of the summary is less than 15\% of that of the original video. 
Kernel-based Temporal Segmentation (KTS) \cite{potapov2014category} is used to segment the videos into shots. The score of a shot is the average score of the frames within it. 

\subsection{Video Summarization with Text Information}

With the pretrained text encoder, we can optionally incorporate text information into video summarization.
As shown in Figure \ref{pvs}(c),
given the text information, the \texttt{[CLS]} token is prepended to the whole sequence, and the pretrained word embedding model is applied to convert the text to a sequence of word embeddings $\bm{X}=\left[\bm{x}_0,\cdots,\bm{x}_{L}\right]^{\mathrm{T}}\in\mathbb{R}^{(L+1)\times {d_w}}$, where $L$ and $d_w$ 
are the length of the word sequence and the dimension of the word embedding. The sequence is then encoded by the pretrained text encoder ($\mathrm{E}_{\mathcal{T}}$),
\begin{equation}
\bm{Z}=\mathrm{E}_{\mathcal{T}}(\bm{X}+\bm{E}^\mathcal{T}_{pos}),
\end{equation}
where $\bm{Z}=\left[\bm{z}_0,\cdots,\bm{z}_{L}\right]^{\mathrm{T}}\in\mathbb{R}^{(L+1)\times {d_w}}$ is the sequence of the encoded word embeddings. $\bm{E}^\mathcal{T}_{pos}\in\mathbb{R}^{(L+1)\times {d_w}}$ is the positional encoding for the word sequence.
Finally, $\bm{z}_0\in\mathbb{R}^{d_w}$, the encoded feature of the \texttt{[CLS]} token, is regarded as the feature of the text information, which is fused into the visual modality for video summarization as follows,
\begin{equation}
\bm{s}^1=\sigma ((\bm{G}^1+\bm{F}^1+\mathrm{MLP}_{\mathcal{T}}(\bm{z}_0))\bm{W}^1+b^1).
\end{equation}
Note that the text representation is used in only the first stage. By this means, the frame-level scores are predicted by taking into consideration not only the visual information but also the associated text information.

\section{Experiments}

We conduct experiments to verify the effectiveness of our method. The experimental settings are first explained. We then compare our method with other video summarization methods and video-text pretraining methods. Finally, we perform ablation studies to demonstrate the impact of our contributions. Additionally, \textbf{the predicted scores are visualized in the supplementary material}.

\subsection{Experimental Settings}

\subsubsection{Datasets for Video Summarization}



We use SumMe \cite{gygli2014creating} and TVSum \cite{song2015tvsum} datasets for video summarization.
For SumMe, we regard their video names in the dataset as \textit{search query} of the text information and the other three types of text information are left empty. For TVSum, the video titles are provided in the dataset, and the 10 categories in TVSum are regarded as \textit{search query} of the text information. Besides, we re-collect the \textit{category} information from YouTube (if not available then left emtpy), and \textit{description} is left empty. Apart from SumMe and TVSum, YouTube \cite{de2011vsumm} and OVP\footnote{Open Video Project: \url{https://open-video.org}} are used for the augmented setting and the transfer setting \cite{zhang2016video,zhou2018deep}. 




\subsubsection{Implementation Details}
\label{id}


\noindent\textbf{Multimodal Self-supervised Pretraining.} 
The videos in YTVT are sub-sampled to 2 FPS to reduce temporal redundancy, which is consistent with SumMe and TVSum.
For the video encoder ($\mathrm{E}_{\mathcal{V}}$) and the Transformer ($\mathrm{T}_{\mathcal{V}}$) in Eq. (\ref{recover1}), the number of heads in the multi-head attention is set to 8. The dimension of the feed-forward network is set to 4,096. To increase the generalization ability, we randomly crop 256 frames from each video for training. The uncased base version of the BERT model is used as the text encoder ($\mathrm{E}_{\mathcal{T}}$). To deal with the inconsistency of text lengths in training, we set the maximum lengths of \textit{category}, \textit{search query}, \textit{title}, and \textit{description} to 3, 3, 10, and 50. For those text sequences shorter than the requirements, they are padded with the \texttt{[PAD]} tokens in the end. The margin $m$ in Eq. (\ref{l2}) is set to $\sqrt{2}$. The windows radius $k$ in Eq. (\ref{recover1}) is set to 4. We set $\alpha=1,\beta=5$ in Eq. (\ref{lssl}). The framework is optimized by Adam \cite{kingma2014adam} with batch size 8 and learning rate $10^{-6}$.

\noindent\textbf{Progressive Video Summarization.} The framework is trained by Adam for 40 epochs with batch size 4 and learning rate $10^{-5}$. The maximum length of the videos is set to 512 frames by random temporal cropping. The videos shorter than the requirement are padded with zeros in the end. When training the model with text information, considering the average length of each type of text, we set the maximum lengths of \textit{category}, \textit{search query}, \textit{title}, and \textit{description} to 1, 3, 10, and 15. Following \cite{he2019unsupervised,jung2019discriminative,zhou2018deep},
five-fold cross-validation is performed. 

\subsubsection{Evaluation Metrics}


\noindent\textbf{F-score.} F-score measures the overlap between the generated video summary and the human summary. Specifically, given the generated video summary $\mathcal{V}_s$ and the human summary $\mathcal{V}_{gt}$, the precision $P$ and recall $R$ are computed as $P=\frac{\left| \mathcal{V}_s\cap \mathcal{V}_{gt}\right| }{\left| \mathcal{V}_s\right|},R=\frac{\left| \mathcal{V}_s\cap \mathcal{V}_{gt}\right| }{\left| \mathcal{V}_{gt}\right|}.$
The F-score $F$ is computed as the harmonic average of $P$ and $R$, i.e., $F=\frac{2PR}{P+R}$.

\noindent\textbf{Rank-based Evaluation.} 
Rank-based evaluation is proposed \cite{otani2019rethinking} to address the limitations of F-score. Specifically, given the predicted frame-level scores and the scores annotated by human, two rank correlation coefficients, Kendall’s $\tau$ and Spearman’s $\rho$, are used as the primary comparison metrics in the experiments. 
For the videos with multiple sets of annotations, the average coefficients are taken as the final results, and the same goes for F-score.


\subsection{Comparisons with the State of the Art}

\subsubsection{Comparisons of Rank Correlation Coefficients}

We compare our methods with the state of the art using the rank correlation coefficients, Spearman’s $\rho$ and Kendall’s $\tau$. 
The results are shown in Table \ref{res2}. 

\begin{table}[tbp]
\caption{The results (Kendall’s $\tau$ and Spearman’s $\rho$) on SumMe and TVSum. 
The methods in the first row are unsupervised methods, while those in the second row are supervised methods. 
}\label{res2}
\center
\begin{tabular}{l|cc|cc}
\hline
\multirow{2}{*}{Methods} & \multicolumn{2}{c|}{SumMe} & \multicolumn{2}{c}{TVSum} \\
 & $\tau$ & $\rho$ & $\tau$ & $\rho$ \\ \hline
SGAN \cite{mahasseni2017unsupervised}~ & --- & --- & 0.024 & 0.032 \\
WS-HRL \cite{chen2019weakly} & --- & --- & 0.078 & 0.116 \\
DRDSN \cite{zhou2018deep} & 0.047 & 0.048 & 0.020 & 0.026 \\
RSGN$_{u}$ \cite{zhao2021reconstructive} & 0.071 & 0.073 & 0.048 & 0.052 \\ \hline
dppLSTM \cite{zhang2016video} & --- & --- & 0.042 & 0.055 \\
CSNet$_{s}$ \cite{jung2019discriminative} & --- & --- & 0.025 & 0.034 \\
GLRPE \cite{jung2020global} & --- & --- & 0.070 & 0.091 \\
SumGraph \cite{park2020sumgraph} &---&---&0.094 &0.138\\
HSA \cite{zhao2018hsa} & 0.064 & 0.066 & 0.082 & 0.088 \\
RSGN \cite{zhao2021reconstructive} & 0.083 & 0.085 & 0.083 & 0.090 \\ 
\hline
SSPVS & \underline{0.178} & \underline{0.240} & \underline{0.177} & \underline{0.233} \\
SSPVS+Text & \textbf{0.192} & \textbf{0.257} & \textbf{0.181} & \textbf{0.238}\\ \hline
\end{tabular}
\end{table}

As shown in Table \ref{res2}, SSPVS outperforms existing methods to a large extent on both datasets, which means the proposed method can model the relative importance among frames more accurately than previous works. Additionally, by including the text information in the summarization process, the performance is further improved on both datasets.

\subsubsection{Comparisons of F-Score}

We also compare our method with previous works in the widely-used F-score. The results are shown in Table \ref{res1}.
Since the text information of the videos in OVP and YouTube is not collected, the results of SSPVS+Text in the augmented setting and the transfer setting are not reported.

As shown in Table \ref{res1}, SSPVS outperforms most of the compared methods, which proves its effectiveness in pinpointing the important shots. We also find that our method is inferior to DSNet \cite{zhu2020dsnet} and MSVA \cite{ghauri2021supervised}. However, DSNet formulates video summarization as temporal detection and requires complex training and testing strategy, which is less applicable. As for MSVA, it uses extra C3D-based features, which is effective but also brings large computation.


\begin{table}[tbp]
\caption{F-score in different settings on SumMe and TVSum. The methods in the first row are unsupervised methods, while those in the second row are supervised methods. \textit{Can}/\textit{Aug}/\textit{Tran} represents the canonical/augmented/transfer setting.}
\label{res1}
\center
\resizebox{0.95\columnwidth}{!}{\begin{tabular}{l|ccc|ccc}
	\hline
	\multirow{2}{*}{Methods} & \multicolumn{3}{c|}{SumMe} & \multicolumn{3}{c}{TVSum} \\
	& \textit{Can} & \textit{Aug} & \textit{Tran} & \textit{Can} & \textit{Aug} & \textit{Tran}\\ \hline
	SGAN \cite{mahasseni2017unsupervised}& 0.387  & 0.417 & ---  &  0.508 & 0.589 & --- \\
	DRDSN \cite{zhou2018deep}& 0.414  & 0.428  & 0.424  & 0.576  & 0.584  & 0.578  \\
	ACGAN \cite{he2019unsupervised}&0.460 & 0.470  &0.445 &0.585 & 0.589 & 0.578 \\
	WS-HRL \cite{chen2019weakly}&0.436 & 0.445  &--- & 0.584 &0.585  & --- \\
	\hline
	vsLSTM \cite{zhang2016video}& 0.376  & 0.416  & 0.407 &0.542 & 0.579  & 0.569 \\
	SGAN$_{s}$ \cite{mahasseni2017unsupervised}& 0.417 & 0.436  &---  &0.563  &0.612  & ---   \\
	H-RNN \cite{zhao2017hierarchical}& 0.421 & 0.438  & --- & 0.579 & \underline{0.619} &  ---  \\
	HSA \cite{zhao2018hsa} & 0.423 & 0.421  & --- & 0.587 & 0.598 &  ---  \\
	re-S2S \cite{zhang2018retrospective} & 0.425 & 0.449  & --- & 0.603 & \textbf{0.639} & ---   \\
	S-FCN \cite{rochan2018video}& 0.475 & \textbf{0.511} & 0.441 & 0.568 &0.592 & \textbf{0.582}  \\
	CSNet$_{s}$ \cite{jung2019discriminative}&0.486 &  0.487 & 0.441&0.585 & 0.571 & 0.574 \\ 
	DSNet \cite{zhu2020dsnet} & 0.502&  \underline{0.507} & \textbf{0.465}&\textbf{0.621}& \textbf{0.639} & \textbf{0.594}\\
    MSVA \cite{ghauri2021supervised} & \textbf{0.534}&  --- & --- &\underline{0.615}&  --- &  ---\\
\hline
	 SSPVS & 0.487  &0.504 & \underline{0.458}  &  0.603& 0.618 &0.578 \\
	 SSPVS+Text & \underline{0.507}  & ---& ---  &  0.604& ---  & --- \\ \hline
\end{tabular}}
\end{table}

\subsubsection{\new{Comparisons of Self-supervised Models}} 
We also compare the proposed self-supervised learning method with other frameworks, including VideoBERT \cite{sun2019videobert}, VideoClip \cite{xu2021videoclip}, and VATT \cite{akbari2021vatt}. 
For fair comparisons, we pretrain the compared models on YTVT with the same form of inputs as our method. Specifically, instead of using images (or their quantizations) as video input, we use the extracted VGG features of frames. Besides, the acoustic modality in VATT is discarded. Then the pretrained video encoders are fine-tuned for video summarization, where the progressive mechanism and text information are not applied. The baseline is our model without pretraining. We report the results of rank-based evaluation in Table \ref{sslres}. As the results show, VATT pretraining reveals little impact on video summarization, while VideoBERT improves the performance marginally on both datasets. Besides, VideoClip improves the results on TVSum significantly.
As for our framework, it improves the results significantly, which indicates the superiority of our method over other self-supervised methods in video summarization. Additionally, \textbf{the impact of each proposed self-supervised loss ($\mathcal{L}_1$, $\mathcal{L}_2$, and $\mathcal{L}_3$) is shown in the supplementary material}.

\begin{table}[tbp]
\caption{The results (Kendall’s $\tau$ and Spearman’s $\rho$) of different self-supervised methods.
}\label{sslres}
\center
\resizebox{0.8\columnwidth}{!}{\begin{tabular}{l|cc|cc}
\hline
\multirow{2}{*}{Methods} & \multicolumn{2}{c|}{SumMe} & \multicolumn{2}{c}{TVSum} \\
 & $\tau$ & $\rho$ & $\tau$ & $\rho$ \\ \hline
Baseline & 0.137 & 0.187 & 0.141 & 0.185 \\\hline
VATT \cite{akbari2021vatt} & 0.137&0.185&0.143&0.188\\ 
VideoBERT \cite{sun2019videobert} & 0.142&0.191&0.145&0.190 \\
VideoClip \cite{xu2021videoclip}&0.139&0.188&0.148&0.195\\
\hline
SSPVS &\textbf{0.154}&\textbf{0.207}&\textbf{0.151}&\textbf{0.199} \\ \hline
\end{tabular}}
\end{table}

\begin{table}[tbp]
\caption{The results of ablation studies in rank-based evaluation.}
\label{abl}
\center
\resizebox{0.9\columnwidth}{!}{\begin{tabular}{ccccccc}
\hline
\multirow{2}{*}{\#Stages} & \multirow{2}{*}{Pretrain} & \multirow{2}{*}{Text} & \multicolumn{2}{c}{SumMe} & \multicolumn{2}{c}{TVSum} \\
 &  &  & $\tau$ & $\rho$ & $\tau$ & $\rho$ \\ \hline
\multirow{3}{*}{1} &  &  & 0.137 & 0.187 & 0.141 & 0.185 \\
 & \checkmark &  & 0.154 & 0.207 & 0.151 & 0.199 \\
 & \checkmark & \checkmark & 0.159 & 0.212 & 0.157 & 0.206 \\ \hline
\multirow{3}{*}{2} &  &  & 0.140 & 0.189 & 0.159 & 0.209 \\
 & \checkmark &  & 0.162 & 0.217 & 0.161 & 0.212 \\
 & \checkmark & \checkmark & 0.174 & 0.235 & 0.163 & 0.214 \\ \hline
\multirow{3}{*}{3} &  &  & 0.166 & 0.224 & 0.162 & 0.212 \\
 & \checkmark &  & \underline{0.178} & \underline{0.240} & 0.172 & 0.226 \\
 & \checkmark & \checkmark & \textbf{0.192} & \textbf{0.257} & 0.173 & 0.228 \\ \hline
 \multirow{3}{*}{4} &   && 0.145&0.198&0.163&0.214 \\
 & \checkmark && 0.172 & 0.231&\underline{0.177}&\underline{0.233} \\
 & \checkmark & \checkmark & 0.175&0.237& \textbf{0.181}&\textbf{0.238}\\ \hline
\end{tabular}}
\end{table}

\subsection{Ablation Study}

We conduct ablation studies to demonstrate the effectiveness of the multimodal self-supervised pretraining, the progressive summarization, and the summarization with text information. Considering the computational complexity and the GPU memory limit, we set the number of stages to 1--4. For each number of stages, three models are evaluated: the base model (without pretraining and text information), the model with pretraining, and the model with pretraining and text information. The results are shown in Table \ref{abl}. 

As the results show, self-supervised pretraining improves the performance significantly for each number of stages. We find that the impact on SumMe is greater than that on TVSum. We believe the reason is that SumMe consists of less training videos than TVSum, and so the pretraining can be more useful on SumMe. Additionally, exploiting text information also benefits video summarization, but the impact becomes less obvious when the number of stages increases.  As for the progressive mechanism, with the increase of the number of stages, the performance is improved on both datasets. The best performance is reached at three stages on SumMe and at four stages on TVSum. 




\new{\section{Conclusion}
We have successfully incorporated video summarization into the self-supervised learning framework which leverages the coarse-grained and fine-grained semantic consistency between the video and the text as well as the temporal dependencies in videos. Based on the pretrained encoders, we have developed progressive video summarization, where the input sequences are refined in a multi-stage fashion and the text information can also be leveraged. Extensive experiments have verified the effectiveness of our contributions. Compared with previous works, our method have achieved state-of-the-art performance in rank-based evaluation.}

\section*{Acknowledgement}
{\small This research was undertaken using the LIEF HPC-GPGPU Facility hosted at the University of Melbourne. This Facility was established with the assistance of LIEF Grant LE170100200. MG was supported by ARC DE210101624.}

\newpage

{\small
\bibliographystyle{ieee_fullname}
\bibliography{egbib}
}

\end{document}